\documentclass[a4paper, 10pt, conference]{ieeeconf}      
\usepackage[a4paper,
            left=54pt,
            right=34pt,
            top=103pt,
            bottom=54pt]{geometry}

\IEEEoverridecommandlockouts                              
\usepackage[utf8]{inputenc}
\usepackage[english]{babel}

\usepackage{cite}
\usepackage{multirow}
\usepackage{graphicx} 
\usepackage[table,xcdraw]{xcolor}
\usepackage{epsfig} 
\usepackage{subfigure}
\usepackage{siunitx}
\usepackage{caption}
\usepackage{bm}
\usepackage{xcolor}

\usepackage{amsmath} 
\usepackage{amssymb}  
\usepackage{amsthm}
\usepackage{mathtools}

\theoremstyle{plain}

\newtheorem{Remark}{Remark}

\usepackage{tabularx}
\usepackage{booktabs}
\usepackage{flafter} 

\usepackage{algorithm}
\usepackage[]{algpseudocode} 

\usepackage{program}

\usepackage{breqn}

\usepackage{siunitx} 

\usepackage{cleveref}

\graphicspath{{figures/}}

\title{\LARGE \bf
Modeling and Control of an Omnidirectional Micro Aerial Vehicle Equipped with a Soft Robotic Arm}

\author{Róbert Szász$^{1}$, Mike Allenspach$^{2}$, Minghao Han$^{1}$, Marco Tognon$^{2}$, Robert K. Katzschmann$^{1}$
\thanks{$^{1}$R. Szász, M. Han and R. Katzschmann are with the Soft Robotics Lab, 
        Swiss Federal Institute of Technology in Zürich (ETH Zürich), Tannenstrasse 3, 8092 Zürich, Switzerland, 
        {\tt\small rszasz@student.ethz.ch, minghao.han@srl.ethz.ch, rkk@ethz.ch}}%
\thanks{$^{2}$M. Allenspach and M. Tognon are with Autonomous Systems Lab, Swiss Federal Institute of Technology in Zürich (ETH Zürich), Leonhardstrasse 21, 8092 Zürich, Switzerland,
        {\tt\small mike.allenspach@mavt.ethz.ch, mtognon@ethz.ch}}%
}

\begin{document}

\maketitle
\thispagestyle{empty}
\pagestyle{empty}


\begin{abstract}
Flying manipulators are aerial drones with attached rigid-bodied robotic arms and belong to the latest and most actively developed research areas in robotics. The rigid nature of these arms often lack compliance, flexibility, and smoothness in movement. This work proposes to use a soft-bodied robotic arm attached to an omnidirectional micro aerial vehicle (OMAV) to leverage the compliant and flexible behavior of the arm, while remaining maneuverable and dynamic thanks to the omnidirectional drone as the floating base. The unification of the arm with the drone poses challenges in the modeling and control of such a combined platform; these challenges are addressed with this work. We propose a unified model for the flying manipulator based on three modeling principles: the Piecewise Constant Curvature (PCC) and Augmented Rigid Body Model (ARBM) hypotheses for modeling soft continuum robots and a floating-base approach borrowed from the traditional rigid-body robotics literature. To demonstrate the validity and usefulness of this parametrisation, a hierarchical model-based feedback controller is implemented. The controller is verified and evaluated in simulation on various dynamical tasks, where the nullspace motions, disturbance recovery, and trajectory tracking capabilities of the platform are examined and validated. 
The soft flying manipulator platform could open new application fields in aerial construction, goods delivery, human assistance, maintenance, and warehouse automation. 

\end{abstract}


\section{Introduction} \label{sec:introduction}

\begin{figure}[t]
	\centering
    \includegraphics[width = .9\columnwidth]{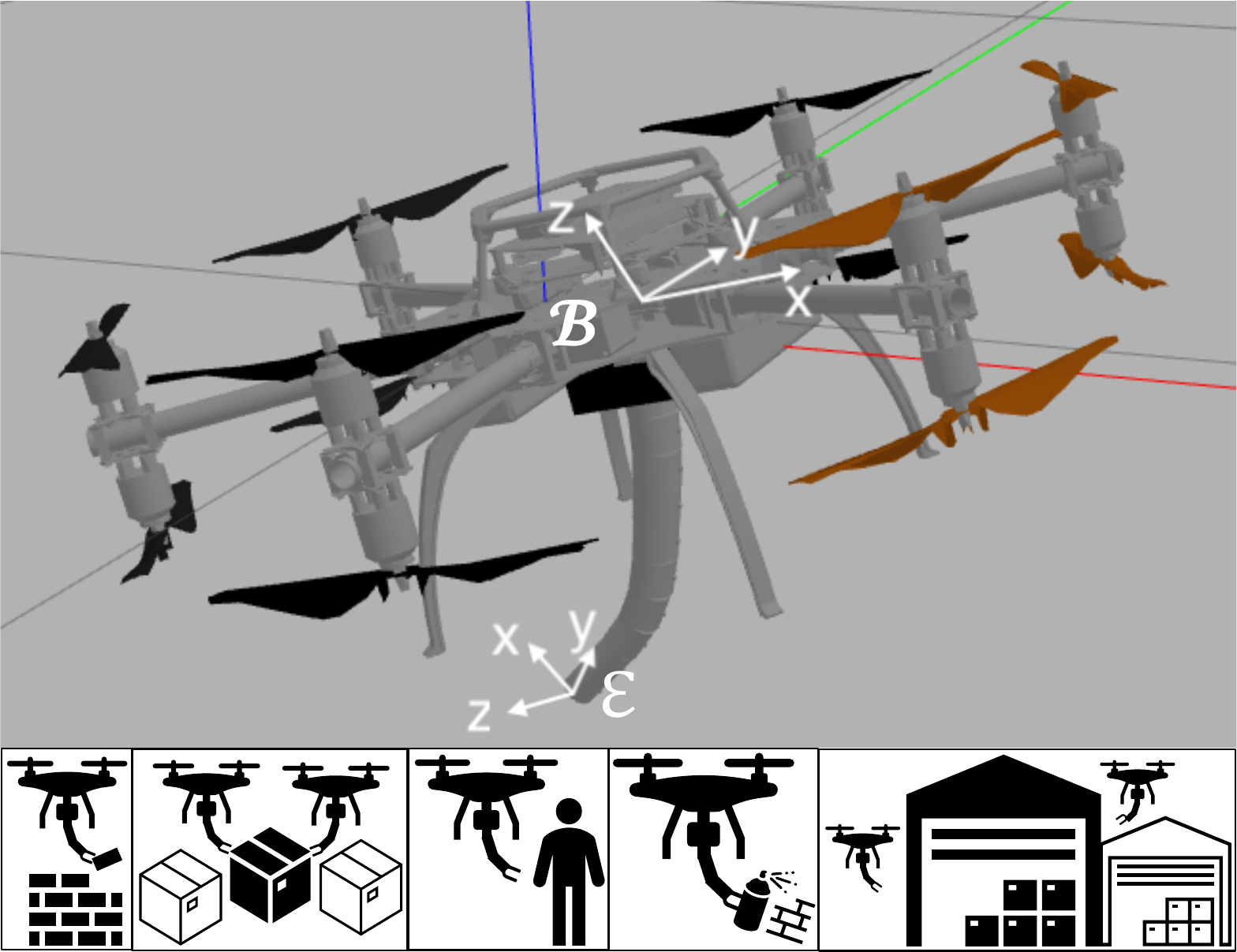}
    \caption{Visualisation of the coupled flying platform consisting of the OMAV and the soft robotic arm used for the controlled simulations. The coordinate frame $\mathcal{B}$ is body-fixed with its origin at OMAV's center of mass and axis aligned with OMAV main inertia axis. The coordinate frame $\mathcal{E}$ is fixed to the tip of the end-effector. When the end-effector is in its fully straight configuration, the frame $\mathcal{E}$ is rotated by 180-degrees around the y-axis with respect to the frame $\mathcal{B}$. The bottom row of images shows some of the possible application areas of such a system: construction, goods delivery, human assistance, maintenance, and warehouse automation.\label{fig:coulped_system}}
    \vspace{-10pt}
\end{figure}

Unmanned aerial vehicles are able to greatly extend the workspace of a robotic arm~\cite{2021-BodTogSie}, which opens up new possibilities in aerial object manipulation and transportation~\cite{2021g-OllTogSuaLeeFra}. However, the major shortcoming of rigid robot arms is their limited capability to interact with humans. The lack of compliance and smooth motions forces these robots to operate in well-structured and fenced-off environments. In regards to the floating base, common quadcopters and hexacopters with fixed propellers are underactuated due to the orientation of their thrusters in the same direction, which makes the modeling and control more complex when using them as the flying base for such aerial manipulators. 
We propose the combination of a soft robotic arm mounted on a fully actuated Omnidirectional Micro Aerial Vehicle (OMAV)~\cite{allenspach_OMAV} (see \Cref{fig:coulped_system}) to tackle these challenges and to counteract the deficiencies of current aerial manipulators.

Robotics researchers have tried in recent years to improve physical human-robot interactions for rigid robots by introducing compliance to their structure, either directly in the joints~\cite{hwisu_compliance_joint}, in the links~\cite{dissertation_compliant_links}, or in the control software~\cite{schumacher2019_compliant_control}. Analogously, compliant joints were proposed for robots mimicking animals~\cite{hutter_starleth_compliance_actuator}. Controlling these advanced rigid-bodied robots in an optimal and robust fashion opens up new challenges and patterns~\cite{hutter_starleth_compliance_control}.

The aforementioned issues and considerations on the lack of inherent compliance gave rise to the emergence of soft robotics. Soft robotics concentrates on building robots that mimic nature as close as possible. As opposed to ``\emph{classical}'' robotics, soft robotics shifts paradigms in the areas of material selection, actuator design, fabrication, and construction. 

A direct counterpart to the conventional rigid robot arms is a controllable soft continuum arm~\cite{katzschmann2015softarm2D}. A follow up work~\cite{katzschmann2020softarm2Dcontrol} extended the notions to model-based feedback control both in parameter-space and Cartesian operational-space, enhancing the capabilities of such soft arms. The soft continuum arms were recently elevated from 2D to 3D using either a model-based feedback control approach based on an Augmented Rigid Body Model (ARBM)~\cite{katzschmann2019softarm3Dcontrol}, or a reduced-order finite element method model based on proper orthogonal decomposition and a state observer~\cite{katzschmann2019softarm3DFEM}.

Similar to classical rigid-bodied systems, the workspace of the soft robotic arm can be extended by mounting it to a flying robot. Instead of choosing a traditional drone, we liberate the soft arm from its fixed base by mounting it to an OMAV for increased dexterity and maneuverability as seen in \Cref{fig:coulped_system}. An OMAV, as opposed to commercially available quadcopters and hexacopters with fixed propeller orientation toward the same direction, is a fully actuated flying system and is capable of exerting forces and torques in any arbitrary direction~\cite{2021f-HamUsaSabStaTogFra}. This feature not only allows the soft continuum arm to move from a confined workspace to the free space -- theoretically reaching any point with any arbitrary orientation in 3D-space -- but also makes the control more robust and agile. The OMAV contributes to the agility and flexibility of such a flying platform and the exchange of the rigid robot arm with a soft arm adds the desired compliance and inherent safety that cannot be achieved otherwise. This platform opens a still unexplored solution in the current scenario of aerial manipulation.

In this paper, we derive a mathematical model from the PCC and ARBM approach and unify it with a floating base robot. Moreover, we propose a hierarchical task-prioritizing controller architecture that enables the user to intuitively define high-level tasks in the operational space. The proposed architecture is able to regulate the system to a fixed point, track a trajectory with a given orientation, and exploit the nullspace of the high-priority tasks to execute additional non-interfering background motions.

Consequently, this work contributes with:
\begin{itemize}
\item A mathematical parametrisation and model of the unified flying platform consisting of the OMAV and the soft robotic arm;
\item A hierarchical task-prioritising controller capable of tracking trajectories while exploiting the nullspace of the higher priority tasks;
\item A validation of the hierarchical controller in various simulations, such as nullspace motion, disturbance recovery and trajectory tracking.
\end{itemize}

%
%
%
\section{Background} 
We give an overview of the relevant modeling approaches for soft continuum manipulators with focus on the PCC and ARBM hypotheses and the state-of-the-art control methods.
We then introduce the aerial robotic research field to motivate our choice in using an OMAV.
%

\subsection{Soft continuum modeling}
\label{sec:softroboticarm}

\begin{figure}
	\centering
    \includegraphics[width = .7\columnwidth]{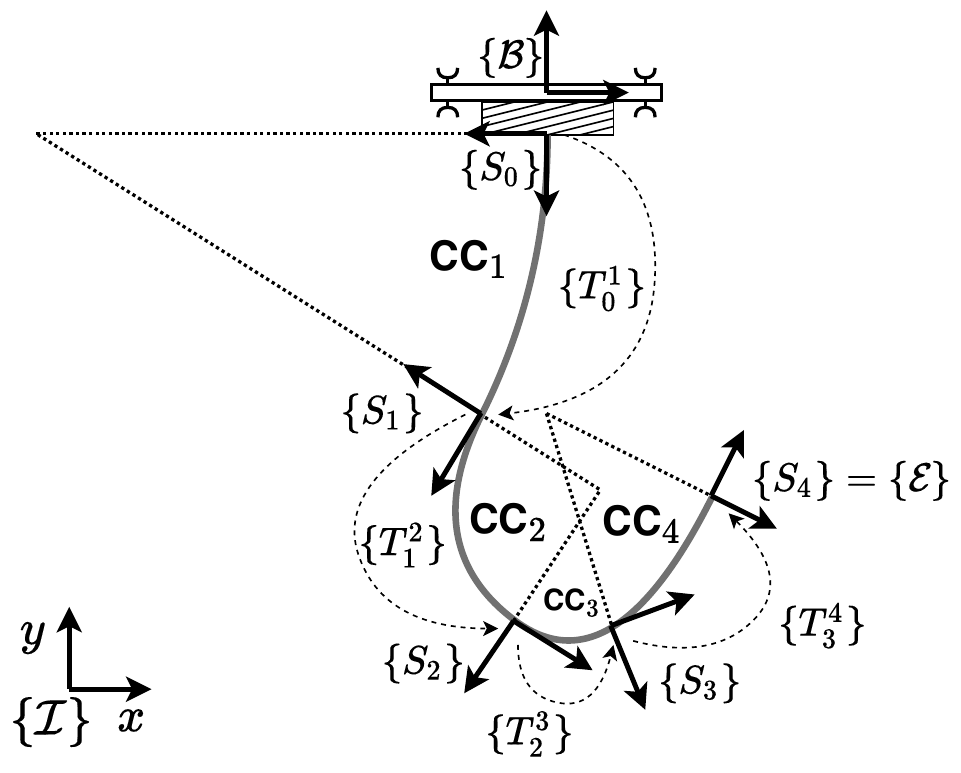}
    \caption{Illustration of the aerial manipulator and the Piecewise Constant Curvature (PCC) arm. The PCC arm is composed of four constant curvature (CC) segments. Frame $\{S_0\}$ denotes the base of the soft robot, while frame $\{S_i\}$ is attached to the end of the current segment and thus the start of the next segment. $T_{i-1}^i$ is the homogeneous transformation mapping from $\{S_{i - 1}\}$ to $\{S_i\}$. The drone's  frame is denoted by $\mathcal{B}$ and the end-effector's  frame is $\{S_4\}=\{\mathcal{E}\}$. Note that the actual system used in the simulation has six CC segments instead of the four presented here.
    \label{fig:PCC_ex}}
\end{figure}

Since soft actuators show high flexibility and compliance, they virtually possess infinite degrees of freedom (DOF), which is challenging from the modeling perspective. Researchers have proposed various techniques for dimensional reduction to render the challenge feasible, for example the \textit{Ritz-Galerkin} models for continuum manipulators~\cite{sadati2018control} and the \textit{Cosserat} approach for soft robots~\cite{renda2017discrete}.

In this paper we use the PCC and ARBM hypotheses. The PCC model approximates the segments of a soft robotic arm with constant bending curvature and treats it as one piece. An example is presented in \Cref{fig:PCC_ex} for a four-way segmented soft arm. 
While PCC is merely a kinematic approximation, previous works \cite{katzschmann2015softarm2D, katzschmann2019softarm3Dcontrol, katzschmann2020softarm2Dcontrol} have confirmed the validity and efficacy of this approach in closed-loop controlled real-world scenarios. The advantage of the PCC and ARBM approach is that modeling approaches and control architectures from the rigid body literature can be directly applied to continuum arms with little effort. 
For the sake of brevity, we do not include here the fundamentals of PCC and ARBM, but we refer the interested reader to~\cite{katzschmann2020softarm2Dcontrol,katzschmann2019softarm3Dcontrol} for a more detailed derivation of this modeling technique.

\subsection{Omnidirectional micro aerial vehicle (OMAV)}
\label{sec:omav}


OMAV aims to eliminate the underactuated system-characteristics thus empowering these flying vehicles with full motion range, six DOF and decoupled controllability between the translational and rotational dynamics.
To obtain such property, different designs have been presented in~\cite{dandrea_omav} and~\cite{ryll_omav}.
In this paper we consider a general tilt-rotor OMAV construction optimized for flight efficiency and large payload capacity, which was introduced and modeled in~\cite{siegwart_OMAV}.

To be able to exploit the abilities of such an agile machine, previous works developed suitable controllers focusing on maneuverability while maintaining flight efficiency at the same time. Decoupling the translation and rotational dynamics while still ensuring robust control made it possible to deploy and use the OMAV for more practical tasks, like contact-based inspections~\cite{siegwart_OMAV_force_control,2019e-TogTelGasSabBicMalLanSanRevCorFra}.

The variable propeller tilting together with the aforementioned controller resulted in an increase in the system's capabilities and maneuverability compared to the fixed-propeller quadrotor or hexarotor. For example, hovering in an arbitrary pose is possible with an OMAV but cannot be achieved with propellers having a fixed tilt angle and facing the same direction (as seen in the classical commercial-grade quad- and hexacopters). Naturally, the modeling and controlling of an OMAV is significantly more challenging than that of a quad- or hexarotor due to the increased capabilities and complexity.

\begin{figure}[t]
    \centering     
    \subfigure[]{\label{fig:pcc_transform}\includegraphics[width=.35\linewidth]{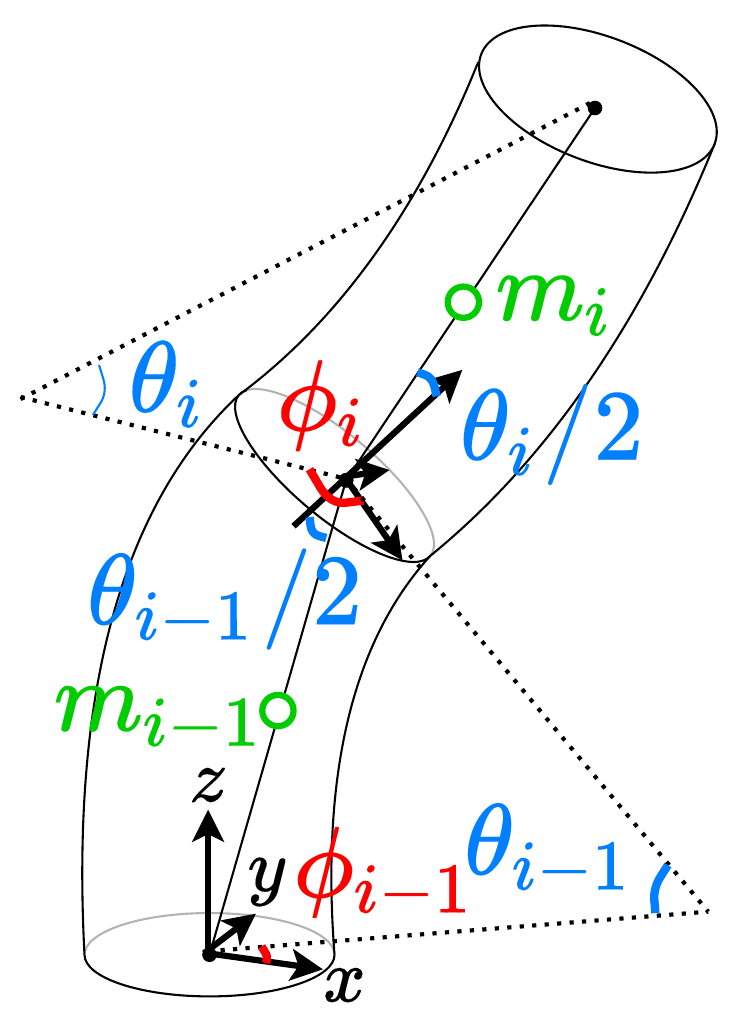}}
    \subfigure[]{\label{fig:augmentation}\includegraphics[width=.55\linewidth]{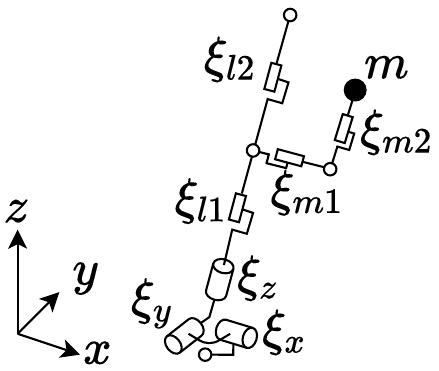}}
    \caption{PCC augmentation and modeling of the segments. (a) Parametrisation and transition between two consecutive curvatures. (b) The extended rigid-body augmentation using 7 joints. Compared to previous works, the two additional joints - $\xi_{m1}$ and $\xi_{m2}$ - account for location of the center of mass off the centerline and thus add an additional degree of matching accuracy. Illustrations are adapted and extended from \cite{yasu2021sopra}.}
\end{figure}

\subsection{Unified flying platform with a manipulator}
\label{sec:flyingplatform}

To the best of our knowledge, there is currently not any work that discusses the control of a flying manipulator consisting of an flying platform and a \emph{soft} robotic arm manipulator. Most research in this field headed towards the direction of controlling a \emph{n}-joint \emph{rigid} body manipulator arm mounted on a \emph{quadcopter}, although most recent works started to explore rigid-body arms mounted on omnidirectional vehicles as well.

Even though we use a soft-robot arm on the flying platform, our hierarchical control architecture is inspired by rigid-arms, such as~\cite{kannan_control_uav_operational_space}, where a 2-joint rigid link arm manipulator was mounted on a drone. The proposed hierarchical control structure consisted of the outermost closed-loop inverse kinematics algorithm layer and a position and attitude control loop inner layer. The results showed that the controller was stable and efficient  with satisfactory performance. \cite{caccavale_control_uav_robot_arm} demonstrated a similar control structure for a 5 DOF arm, further analysing and proving the stability mathematically. A more closely related work is \cite{fishman2021softdrone}, a quadrotor equipped with a soft tendon-driven grasping mechanism attached to it. The control is conducted with an optimisation-based approach consisting of two submodules, separately optimizing for the quadrotor trajectory and the soft gripper movement.

 Our aim is to combine and extend the previous approaches: we build upon the novel tiltrotor OMAV architecture and combine it with the soft continuum arm, we find a PCC/ARBM-supported unified parametrization and we design an analytic model-based hierarchical feedback controller. 


%
%
\section{Model} 
The model shall leverage all six DOF of the OMAV. Therefore, we propose a unified extended floating-base parametrisation for the coupled system. This choice is analogous to the frequently used rigid-body counterpart, as described for example in \cite{nakanishi2007floatingbase}. 

The soft continuum arm is modeled using the PCC and ARBM hypotheses. These hypotheses are based on the assumptions of non-extensible curvature segments and constant curvature along each one of these segments. The PCC approximation offers a suitable one-to-one mapping between the kinematic properties of non-extensible and constant curvature segments and a real-world soft continuum arm segment. The ARBM approach the real-world arm dynamics by describing an \emph{augmented} rigid-robot space with rigid links connecting translational and rotational joints~\cite{katzschmann2018softarm2Dcontrol}. Each segment has its own set of parameters $\theta$ for the bending angle and $\phi$ for the off-plane rotation in the PCC space (see 
\Cref{fig:pcc_transform}). To match the kinematic and dynamic profile of the soft continuum arm, a suitable rigid-body augmentation is required. A trade-off between computational costs resulting from model complexity and modeling accuracy puts limitations on the potential configuration of the chosen rigid-body robot acting dynamically equivalent to the soft arm. 

In this paper, we propose a 7-joint rigid robot configuration, which builds upon and extends the 5-joint parametrisation presented in \cite{yasu2021sopra}. Without $\xi_{m1}$ and $\xi_{m2}$, the mass would have to lie on the virtual line connecting the start and end point of a PCC segment. Adding these two extra joints helps to extend the system's capability by more accurately representing the shift of the center of mass of the soft arm during bending (see \Cref{fig:augmentation}).

\begin{figure}[t]
	\centering
    \includegraphics[width = .65\columnwidth]{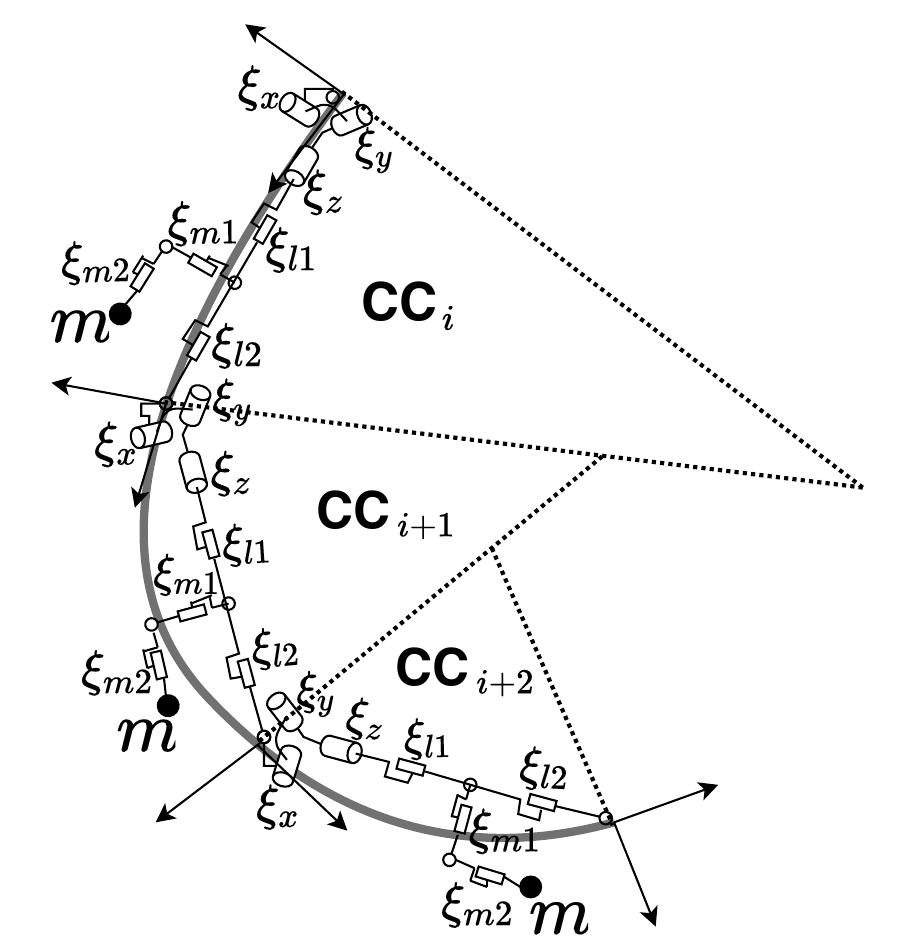}
    \caption{Three PCC segments overlaid with their augmentations.  \label{fig:augmentation_full}}
\end{figure}

In previous work~\cite{katzschmann2020softarm2Dcontrol,katzschmann2019softarm3Dcontrol}, one PCC element was mapped to one actuated segment of the soft continuum arm. However, we decided to take a more recent approach~\cite{yasu2021sopra} and increase the fidelity of simulation by modeling the actuated segments $N_{seg}=2$ each with $N_{PCC}=3$ PCC elements, summing up to a total of $N_{seg}*N_{PCC}*N_{aug}=2*3*7 = 42$ joints. The augmentation of one actuated segment is shown in \Cref{fig:augmentation_full}. This step increases the accuracy of the model at the cost of higher computational demands. Note that the variables $N_{seg}$, $N_{PCC}$, and $N_{aug}$ determine the dimensions of the \emph{augmented} space.

The augmented soft continuum arm is extended with a floating base, \emph{i.e.}, the OMAV. The OMAV is modeled as a mass point with given mass and inertia properties. To represent rotations without a singularity, we use the quaternion elements $\xi_{quat_i}, i \in \{w,x,y,z\}$. The state of the system results in the following parametrisation vector:
\begin{equation}\label{eq:parametrisation}
    \bm{\xi}=
    \begin{bmatrix}
        \bm{\xi}_{base}\\
        \xi_{x_1}\\
        \xi_{y_1}\\
        ... \\
        \xi_{m2_6}
    \end{bmatrix}=
    \begin{bmatrix}
        \bm{\xi}_{quat_{OMAV}}\\
        \bm{\xi}_{pos_{OMAV}}\\
        \xi_{x_1}\\
        \xi_{y_1}\\
        ... \\
        \xi_{m2_6}
    \end{bmatrix}
    \bm{\dot{\xi}}=
    \begin{bmatrix}
        \prescript{}{\mathcal{B}}{\bm{\omega}_{IB}} \\
        \prescript{}{\mathcal{B}}{\bm{v}} \\
        \dot{\xi}_{x_1}\\
        \dot{\xi}_{y_1}\\
        ... \\
        \dot{\xi}_{m2_6}
    \end{bmatrix}
    \bm{\ddot{\xi}}=
    \begin{bmatrix}
        \prescript{}{\mathcal{B}}{\bm{\dot{\omega}}_{IB}} \\
        \prescript{}{\mathcal{B}}{\bm{a}} \\
        \ddot{\xi}_{x_1} \\
        \ddot{\xi}_{y_1} \\
        ... \\
        \ddot{\xi}_{m2_6}
    \end{bmatrix} \; ,
\end{equation}
where $\bm{\xi}_{pos_{OMAV}} \in \mathbb{R}^{3}$ represents the Cartesian coordinates of the OMAV's centre of mass, while $\bm{v} \in \mathbb{R}^{3}$ and $\bm{a} \in \mathbb{R}^{3}$ are the first and second derivative of this quantity with respect to time. The vector $[\xi_{x_i}, \xi_{y_i}, \xi_{z_i}, \xi_{{l1}_i}, \xi_{{l2}_i}, \xi_{{m1}_i}, \xi_{{m2}_i}]^T \in \mathbb{R}^7, i \in {1...6}$ denotes the augmentation of the constant curvature segments, as depicted in \Cref{fig:augmentation}.

The calligraphic prescripts -- $\mathcal{B}$ refers to the body-fixed coordinate system attached to the center of mass of the OMAV and axis aligned with main axis of inertia, and $\mathcal{I}$ is the inertial coordinate frame -- denotes the coordinate system, in which the quantity is described. The double subscript in $\bm{\omega}_{IB} \in \mathbb{R}^{3}$ denotes the angular velocity of the $\mathcal{B}$ coordinate frame with respect to the inertial frame $\mathcal{I}$ and $\bm{\dot{\omega}}_{IB} \in \mathbb{R}^{3}$ represents the angular acceleration. 
Note in \eqref{eq:parametrisation} that the parametrisation vector  $\bm{\xi} \in \mathbb{R}^{N_{seg}*N_{PCC}*7+7}=\mathbb{R}^{49}$, and  the first and second derivative terms are from the vector space $\bm{\dot{\xi}},\bm{\ddot{\xi}} \in \mathbb{R}^{48}$. This is due to the choice of a singularity-free rotation representation using quaternions with 4 elements. 

Transforming the hybrid platform to the augmented rigid-robot space enables a fast and straightforward derivation of the Jacobian matrix for an arbitrary fixed-point $Q$ in the body coordinate system:
\begin{equation}\label{eq:jacobian}
\begin{split}
    & \prescript{}{\mathcal{I}}{\bm{r}}_{IQ}=\prescript{}{\mathcal{I}}{\bm{r}}_{IB}+\bm{C}_{IB}\prescript{}{\mathcal{B}}{\bm{r}}_{BQ} \\
	& \prescript{}{\mathcal{I}}{\bm{v}}_{Q}=\prescript{}{\mathcal{I}}{\bm{v}}_{B}+\dot{\bm{C}}_{IB}\prescript{}{\mathcal{B}}{\bm{r}}_{BQ}+\bm{C}_{IB}\prescript{}{\mathcal{B}}{\dot{\bm{r}}}_{BQ} \\
	& \prescript{}{\mathcal{I}}{\bm{v}}_{Q}=\bm{C}_{IB}\prescript{}{\mathcal{B}}{\bm{v}}_{B}-\bm{C}_{IB}\begin{bsmallmatrix}\prescript{}{\mathcal{B}}{\bm{r}}_{BQ} \end{bsmallmatrix}_{\times}\prescript{}{\mathcal{B}}{\bm{\omega}}_{IB}+\bm{C}_{IB}\prescript{}{\mathcal{B}}{\bm{J}}_{Q}\dot{\xi}_j
\end{split}
\raisetag{2\normalbaselineskip}
\end{equation}
where ${\bm{r}}_{IB} \in \mathbb{R}^{3}$ denotes a vector from the origin $I$ of the inertial frame to the origin of the floating-base body-fixed frame $B$, $\prescript{}{\mathcal{B}}{\bm{J}}_{Q} \in \mathbb{R}^p$ is the local Jacobian written in the body-fixed coordinate frame, and $\bm{\dot{\xi}}_j \in \mathbb{R}^p$ is the time derivative of the robot parametrisation vector without the floating-base ($p$ denotes the number of joints).

For the sake of brevity, the dependency of the terms on $\bm{\xi}$ was omitted. From the last equation, the Jacobian is $\prescript{}{\mathcal{I}}{\bm{J}}_Q(\xi)=
    \begin{bsmallmatrix}
        \bm{C}_{IB}(\xi) & 
        -\bm{C}_{IB}(\xi)\begin{bsmallmatrix}\prescript{}{\mathcal{B}}{\bm{r}}_{BQ}(\xi) \end{bsmallmatrix}_{\times} &
        \bm{C}_{IB}(\bm{\xi})\prescript{}{\mathcal{B}}{\bm{J}}_{Q}
    \end{bsmallmatrix}$, 
where $\bm{C}_{IB} \in \mathbb{R}^{3 \times 3}$ is the rotation matrix of the body-frame with respect to the inertial frame and $\begin{bsmallmatrix}\bm{u}\end{bsmallmatrix}_{\times}$  denotes the cross-product skew symmetric matrix created from the vector $\bm{u}$. With the Jacobian, the inertia matrix $\bm{B}_{\xi}(\bm{\xi}) \in \mathbb{R}^{(N_{seg}*N_{PCC}*7 + 6) \times (N_{seg}*N_{PCC}*7 + 6)}=\mathbb{R}^{48 \times 48}$, Coriolis and centrifugal vector $\bm{c}_{\xi}(\bm{\dot{\xi}}, \bm{\xi}) \in \mathbb{R}^{N_{seg}*N_{PCC}*7 + 6}=\mathbb{R}^{48}$, and the gravitational field vector $\bm{g}_{\xi}(\bm{\xi}) \in \mathbb{R}^{N_{seg}*N_{PCC}*7 + 6}=\mathbb{R}^{48}$ can be derived in the augmented formulation using the equations (3.43), (3.44) and (3.45) from~\cite{robot_dynamics_lecture_notes_eth}. 
%

\begin{Remark}
Instead of parameterizing the constant curvature segments with the off-plane rotation $\phi$ and bending angles $\theta$ as shown in \Cref{fig:pcc_transform}, an alternative parameterization composed of $\theta_x, \theta_y$ taken from \cite{yasu2021sopra} is applied:
\begin{equation}\label{eq:theta_theta_parametrisation}
\begin{split}
    \theta_x := \theta \cos{\phi} \\
    \theta_y := \theta \sin{\phi}
\end{split}
\end{equation}
which avoids a singularity when the soft continuum arm is in its straight configuration. This parametrization is used later for the state vector $\bm{q}$.
\end{Remark}

The bridging between the augmented rigid-body space and the PCC space is described with the set of equations (11) from~\cite{katzschmann2019softarm3Dcontrol}:
\begin{equation} \label{eq:xi}
	\begin{cases}
		\bm{\xi} &= m(\bm{q}) \\
		\bm{\dot{\xi}} &= \bm{J}_{\mathrm{m}}(\bm{q}) \bm{\dot{q}} \\
		\bm{\ddot{\xi}} &= \bm{\dot{J}}_{\mathrm{m}}(\bm{q},\bm{\dot{q}}) \bm{\dot{q}} + \bm{J}_{\mathrm{m}}(\bm{q}) \bm{\ddot{q}}
	\end{cases}
\end{equation}
where $\bm{q} \in \mathbb{R}^{19}$ is the coupled system's parametrisation in the PCC space consisting of the $\theta_x, \theta_y$ parameters for the $N_{seg}*N_{PCC}=2*3=6$ PCC segments, the 4 quaternion elements, and the 3 Cartesian coordinates of the floating base. $m(\cdot): \mathbb{R}^{19} \rightarrow \mathbb{R}^{49}$ maps the PCC parametrisation to the augmented space and $\bm{J}_{\mathrm{m}}(\bm{q}) \in \mathbb{R}^{48 \times 18}$ is the Jacobian of $m(\cdot)$, \emph{i.e.}, $\frac{\partial m}{\partial \bm{q}}$. Its upper-left $6 \times 6$ sub-matrix affecting the OMAV parametrisation from one space to the other is  the identity matrix, since the mapping is one-to-one.

Using the mapping above, the PCC and ARBM assumptions \cite{katzschmann2019softarm3Dcontrol} define the system-matrices as follows:
\begin{equation}\label{eq:system_matrices}
	\begin{cases}
		\bm{B}(\bm{q}) &= \bm{J}_{\mathrm{m}}^T(\bm{q}) \, \bm{B}_{\xi}(m(\bm{q})) \, \bm{J}_{\mathrm{m}}(\bm{q}) \\
		\bm{c}(\bm{q},\bm{\dot{q}}) &= \bm{J}_{\mathrm{m}}^T(\bm{q}) \, \bm{c}_{\xi}(m(\bm{q}), \bm{J}_{\mathrm{m}}(\bm{q})\bm{\dot{q}}) \\
		\bm{g}(\bm{q}) &= \bm{J}_{\mathrm{m}}^T(\bm{q}) \, \bm{g}_{\xi}(m(\bm{q}))\\
        \bm{J}(\bm{q}) &= \bm{J}_{\xi}(m(\bm{q})) \, 
        \bm{J}_{\mathrm{m}}(\bm{q})
	\end{cases}
\end{equation}
where $\bm{B}_{\xi}, \bm{c}_{\xi}, \bm{g}_{\bm{\xi}}$ and $\bm{J}_{\xi}$ are augmented space quantities and have dimensions as described above. Finally, the dynamics of the coupled system are given as (for the full derivation please refer to \cite{katzschmann2018softarm2Dcontrol}):
\begin{equation}\label{eq:soft_robot_dynamics}
	\begin{split}
		\bm{B}(\bm{q})\bm{\ddot{q}} + \bm{c}(\bm{q}, \bm{\dot{q}}) + \bm{g}(\bm{q}) + \bm{K} \bm{\Tilde{q}} + \bm{D} \bm{\dot{q}} = \\ 
		\bm{\Tilde{A}}_{\alpha} \bm{\Omega} + \bm{S}_{\mathrm{sel}} \bm{A} \bm{p} + \bm{J}^T(\bm{q}) \bm{f}_{\mathrm{ext}}
	\end{split}
\end{equation}
which is analogous to the standard rigid-body formulation. With $n=18$, $\bm{B}(\bm{q}) \in \mathbb{R}^{n \times n}$ is the inertia matrix, $\bm{c}(\bm{q}, \bm{\dot{q}}) \in \mathbb{R}^{n}$ is the vector containing the Coriolis and centrifugal terms, $\bm{g}(\bm{q})$ is the gravitational force equivalent. The stiffness of the soft continuum arm is considered in the stiffness matrix $\bm{K} \in \mathbb{R}^{n \times n}$ and the damping effects are expressed by the damping matrix $\bm{D} \in \mathbb{R}^{n \times n}$. For the exact derivation and calculation of the stiffness and damping, please refer to \cite{yasu2021sopra} section \emph{III. C}. Both matrices have an upper-left $6 \times 6$ block of zeros to account for the lack of stiffness or damping in the equations of motion of the OMAV. Note that the stiffness matrix is multiplied with a modified state vector $\bm{\Tilde{q}} \in \mathbb{R}^{18}$. Since the first 6 elements of the vector are affected by the zero block of the stiffness matrix, the values in the modified state vector are irrelevant and thus set to zero, \emph{i.e.}, $[\mathrm{0}_{1 \times 6}, \theta_{x_1}, \theta_{y_1}, \theta_{x_2}, \theta_{y_2} \dotsc]^T$. This multiplication is necessary because the dimensions of the matrix multiplication would be inconsistent otherwise.  $\bm{J}^T(\bm{q})$ is responsible for mapping external forces $\bm{f}_{\mathrm{ext}}$ from the operational space to the joint space.

The OMAV rotor force generation and the soft robotic arm pressure actuation is on the right hand side of \eqref{eq:soft_robot_dynamics}. The expanded expression for the modified allocation matrix is $\bm{\Tilde{A}}_{\alpha} := \bm{J}_{\mathrm{m}}^T \, \bm{J}_{\mathrm{S}}^T \, \bm{A}_{\alpha} \in \mathbb{R}^{n \times w}$, where $w$ is the number of rotors. The squared rotor speeds $\bm{\Omega} \in \mathbb{R}^w$ multiplied with the instantaneous allocation matrix $\bm{A}_{\alpha}$ results in the body forces and torques, as described by (7) in \cite{allenspach_OMAV}. The wrench is mapped to the augmented rigid-body space with the OMAV center of mass Jacobian $\bm{J}_{\mathrm{S}} \in \mathbb{R}^{6 \times (N_{seg}*N_{PCC}*7 + 6)}$. The last step is the transition from the augmented space to the PCC space, which is carried out using the space mapping Jacobian $\bm{J}_{\mathrm{m}}$. The actuation inclusion of the soft continuum arm in the mathematical description follows analogously. $\bm{A} \in \mathbb{R}^{(n - 6) \times 6}$ is the conversion matrix between the chamber pressures and generalized forces, which acts on the pressure input $\bm{p} \in \mathbb{R}^6$ and is derived in \cite{yasu2021sopra}. Since the pressurisation of the soft arm chamber disregards the OMAV's equation of motions, a selection matrix $\bm{S}_{\mathrm{sel}} \in \mathbb{R}^{n \times (n - 6)}$ is required to transform the input vector to the correct dimensions. To keep the mathematical consistency of the derivation, $\bm{S}_{\mathrm{sel}}$ is composed of two blocks: the upper $6 \times (18 - 6)$ region affecting the OMAV's DOFs are zero, while the lower $(18 - 6) \times (18 - 6)$ is the identity matrix.

\section{Control}
\label{sec:control}

This section introduces our proposed hierarchical control architecture designed for the coupled system. 
The controller is a prioritisation-based hierarchical controller and operates on the end-effector orientation, position, and OMAV orientation tasks. The OMAV position is not taken into account, since the only important position quantity is that of the end-effector. Nevertheless, OMAV orientation can help with avoiding singularities and inefficient configurations. The more important targets from the user's perspective are 
assigned the highest priority and the ordering in priority is as follows:
\begin{enumerate}
  \item \textbf{End-effector orientation:} determined by the offset term $\Delta\bm{\phi}_{EE} \in \mathbb{R}^3$ that depends on the reference orientation $\bm{\phi}_{EE_{d}} \in \mathbb{R}^3$  and the controlled orientation $\bm{\phi}_{EE} \in \mathbb{R}^3$, both represented as angle-axis.
  \item \textbf{End-effector position:} $\Delta \bm{x}_{EE} \in \mathbb{R}^3$ is the difference of reference Cartesian position $\bm{x}_d \in \mathbb{R}^3$ and the current Cartesian position $\bm{x} \in \mathbb{R}^3$ of the end-effector.
  \item \textbf{OMAV orientation:} separately controlled thanks to the high number of DOFs and the hierarchical prioritisation approach. $\Delta\bm{\phi}_{O} \in \mathbb{R}^3$ depends on the reference OMAV rotation $\bm{\phi}_{O_{d}} \in \mathbb{R}^3$ and current orientation $\bm{\phi}_O \in \mathbb{R}^3$. Both rotations are represented in angle-axis.
\end{enumerate}
\begin{figure}
	\centering
    \includegraphics[width = 1.0\columnwidth]{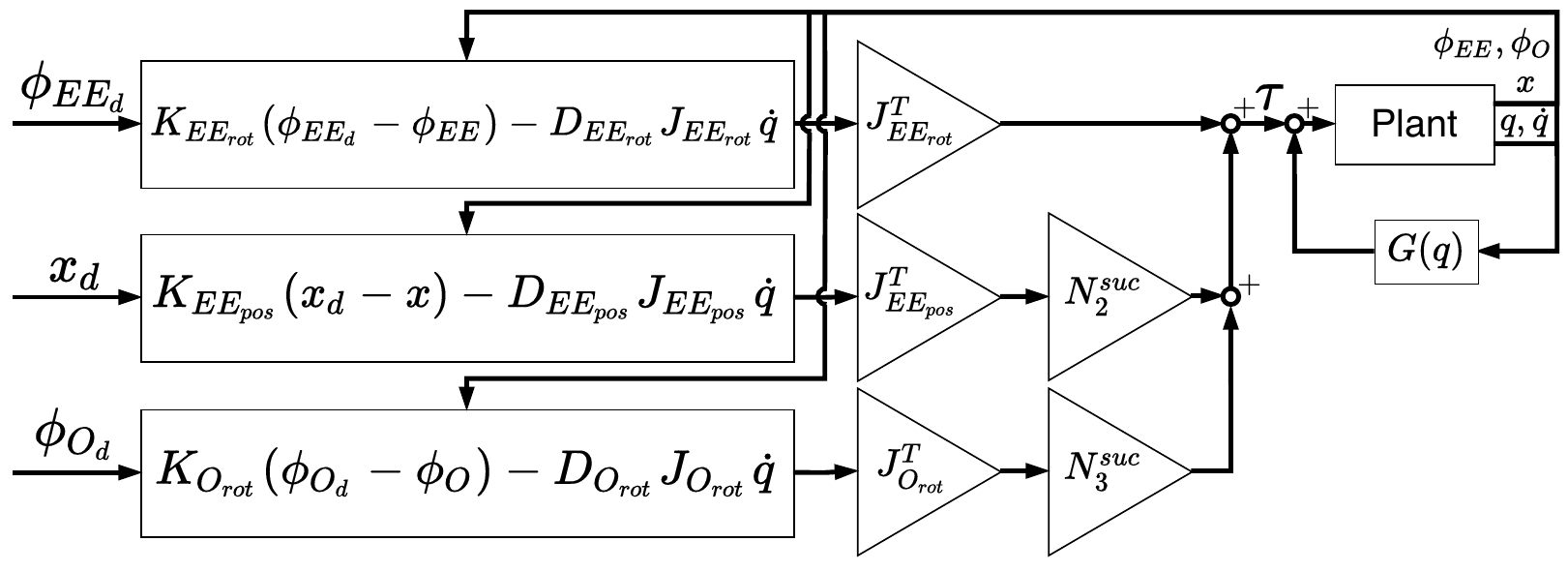}
    \caption{Block diagram of the hierarchical controller. The nullspace projection matrices $N_2^{suc}$ and $N_3^{suc}$ prevent the lower priority tasks from interfering with higher priority tasks and ensure consistency not just in steady state, but also in the transient phase. \label{fig:blockdiagram}}
\end{figure}

The angular offset $\Delta\bm{\phi}$ is calculated using the definitions from \cite{robot_dynamics_lecture_notes_eth}. The current orientation $\mathcal{S} \in {\mathcal{E}, \mathcal{B}}$ (end-effector and OMAV frame) is rotated back to an arbitrary inertial frame $\mathcal{I}$, then rotated to the goal frame $\mathcal{G}$ using the rotation equation and the orthonormality of rotation matrices:
\begin{equation} \label{eq:rotation}
	\bm{C}_{\mathcal{G}\mathcal{S}}(\Delta\bm{\phi}) = \bm{C}_{\mathcal{G}\mathcal{I}}(\bm{\phi}_d) \bm{C}_{\mathcal{S}\mathcal{I}}^T(\bm{\phi}) \; .
\end{equation}
Consequently, the rotation matrix is converted to an angle-axis representation to describe an offset vector in $\mathbb{R}^3$.

The backbone of each task is an offset-driven feedback term presented on the left half of \Cref{fig:blockdiagram}: 
\begin{equation}\label{eq:joint_space_control_2}
\begin{split}
    \bm{\tau}_1 &= \bm{J}_{EE_{rot}}^T (\bm{K}_{EE_{rot}} \Delta \bm{\phi}_{EE} - \bm{D}_{EE_{rot}} \bm{J}_{EE_{rot}} \bm{\dot{q}}) \\
    \bm{\tau}_2 &= \bm{N}_2^{suc} \bm{J}_{EE_{pos}}^T (\bm{K}_{EE_{pos}} \Delta \bm{x}_{EE} - \bm{D}_{EE_{pos}} \bm{J}_{EE_{pos}} \bm{\dot{q}}) \\
    \bm{\tau}_3 &= \bm{N}_3^{suc} \bm{J}_{O_{rot}}^T (\bm{K}_{O_{rot}} \Delta \bm{\phi}_{O} - \bm{D}_{O_{rot}} \bm{J}_{O_{rot}} \bm{\dot{q}}) \; ,
\end{split}
\raisetag{2\normalbaselineskip}
\end{equation}
For the sake of brevity, the matrix dependencies on the full system parametrization vector $\bm{q}$ and its derivative $\bm{\dot{q}}$ were omitted. $\bm{K}_{EE_{rot}}, \bm{D}_{EE_{rot}}, \bm{K}_{EE_{pos}}, \bm{D}_{EE_{pos}}, \bm{K}_{O_{rot}}, \bm{D}_{O_{rot}} \in \mathbb{R}^{3 \times 3}$ are the stiffness and damping tuning matrices for the individual tasks. $\bm{J}_{EE_{rot}}, \bm{J}_{EE_{pos}}, \bm{J}_{O_{rot}} \in \mathbb{R}^{3 \times n}$ are the end-effector rotational, translational, and OMAV rotational Jacobians in the PCC space, respectively.

To simultaneously ensure a task prioritisation and dynamic consistency, successive nullspace projection matrices $\bm{N}_2^{suc}, \bm{N}_3^{suc} \in \mathbb{R}^{n \times n}$ are applied to the tasks, as described in~\cite{dietrich2015nullspace_projection}. A nullspace projector for a task with the Jacobian $\bm{J}$ of a previous task (\emph{e.g.}, the end-effector orientation task $\bm{J}_{EE_{rot}}^T(\bm{K}_{EE_{rot}} \Delta \bm{\phi}_{EE} - \bm{D}_{EE_{rot}} \bm{J}_{EE_{rot}} \bm{\dot{q}})$ given by the controller) is obtained by: 

\begin{equation} \label{eq:nullspace}
	\bm{N}(\bm{q}) = \mathbb{\bm{I}}_{n \times n} - \bm{J}(\bm{q})^T \bm{J}(\bm{q})^{\#^T} \; ,
\end{equation}
where $\bm{J}(\bm{q})^{\#}$ is the generalized inverse satisfying the criterion $\bm{J}(\bm{q}) \bm{J}(\bm{q})^{\#}=\mathbb{\bm{I}}_{n \times n}$. To fulfill the criteria that lower priority tasks not interfere with higher priority tasks during the transient phase or the steady state, a dynamically consistent inverse weighted by the inertia matrix is adapted and denoted as:
\begin{equation} \label{eq:pseudoinverse}
	\bm{J}^{\#}:=\bm{J}^{B+} = \bm{B}^{-1} \bm{J}^T (\bm{J} \bm{B}^{-1} \bm{J}^T)^{-1} \; ,
\end{equation}
based on \cite{dietrich2015nullspace_projection}.

The torque $\bm{\tau}$ calculated by the controller is directly fed into the plant. To avoid a steady state tracking error, a gravitational decoupling was added to the control scheme. For a more detailed description of the output allocation pipeline for the OMAV and for the soft continuum arm, please refer to \cite{allenspach_OMAV} and \cite{yasu2021sopra}, respectively.

\section{Simulations}
The simulation scenarios and tasks were designed to show some key capabilities of the coupled system, like continuous nullspace exploitation, disturbance rejection, or dynamic trajectory tracking. We believe that these simulations show the potential of a coupled soft robot arm and aerial drone system. 
In this section, we first describe the hardware and software setup together with the parameter values used for the simulations. Afterwards, simulation results are shown and discussed.

\subsection{Simulation setup}
The simulations were run on a laptop with a 4-core CPU (Intel i7-6820HQ (4 x 2.7GHz) 6. Generation) and 16GB RAM memory. The operating system used was an Ubuntu 18.04 Bionic.

The code was written in \emph{C++ 17}, with \emph{cmake} version 3.20.0. The visualisation pipeline was built on \emph{ROS 1 - Melodic Morenia}~\cite{ros} and~\emph{Gazebo}, an open source 3D robotics simulator (here only used for visualisation). The \emph{Drake}~\cite{drake} library runs the physics engine in the background for the augmented rigid-body robot system descriptions and simulations. The virtual control frequency was set to 100Hz (this sets an achievable target for the real platform as well) and the internal state update frequency of the system was 100kHz. Note that due to the offline nature of the simulation, no real-time performance was targeted and the code is therefore not necessarily performance-optimised. 

\begin{table}[htp]
\centering
\footnotesize
\begin{tabular}{lcc}
\toprule
& \multicolumn{2}{c}{Control task}\\ \cmidrule(lr){2-3}
 & static & dynamic \\
\midrule
$\bm{K}_{EE_{rot}}$ & $(1.5,1.5,1.5)$  & $(1.5,1.5,1.5)$ \\
$\bm{D}_{EE_{rot}}$ & $(0.025,0.025,0.025)$  & $(0.025,0.025,0.025)$ \\
$\bm{K}_{EE_{pos}}$ & $(13.8,13.8,13.8)$  & $(3.8,3.8,3.8)$ \\
$\bm{D}_{EE_{pos}}$ & $(12.5,12.5,12.5)$  & $(2.5,2.5,2.5)$ \\
$\bm{K}_{O_{rot}}$ & $(5.0,5.0,5.0)$  & $(5.0,5.0,5.0)$ \\
$\bm{D}_{O_{rot}}$ & $(4.0,4.0,4.0)$  & $(4.0,4.0,4.0)$ \\
\bottomrule
\end{tabular}
\caption{Values of the gain and damping matrices}
\label{table:matrices}
\end{table}

\begin{figure}
	\centering
    \subfigure[End-effector position deviation from reference]{\includegraphics[width = .85\columnwidth]{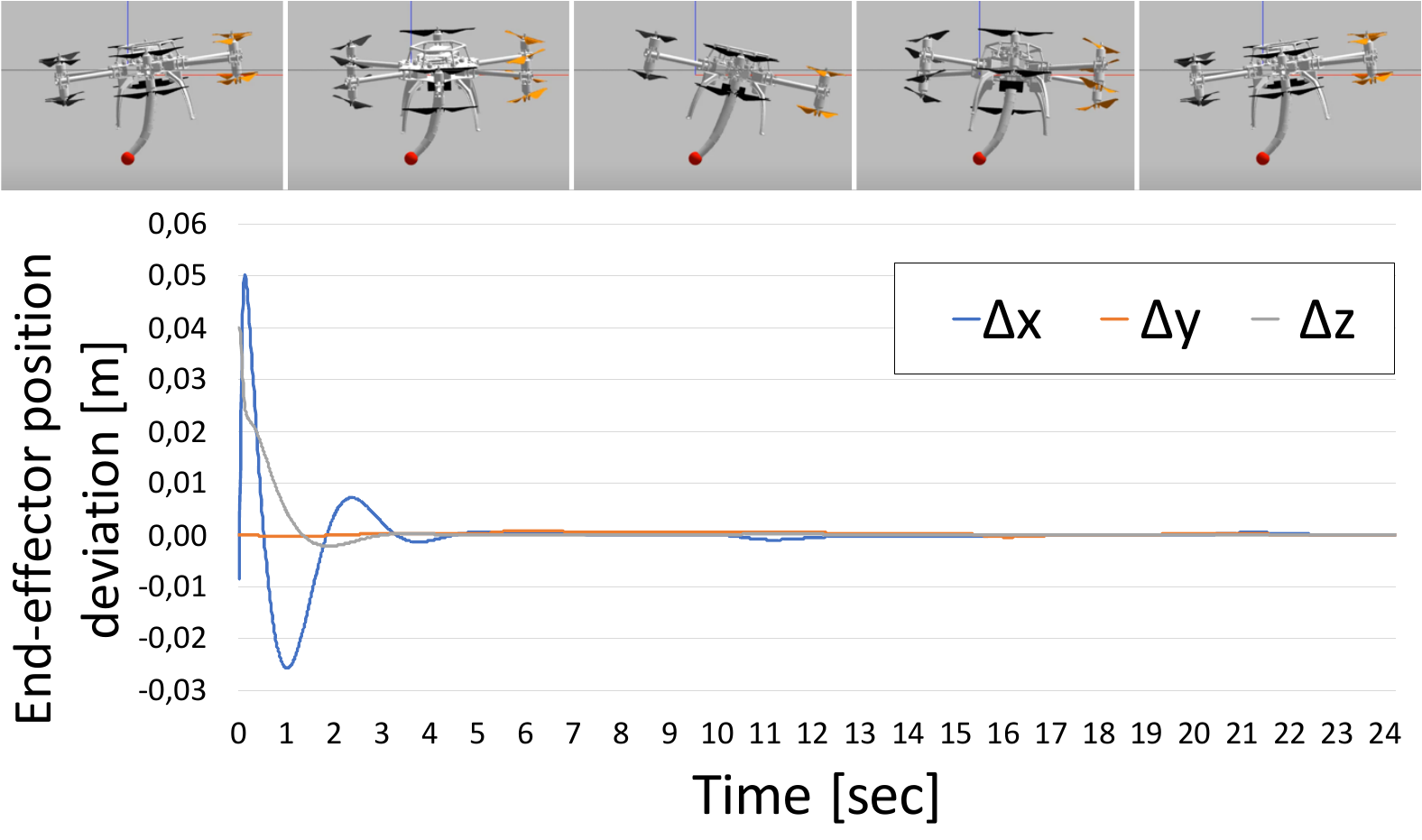}}
    \subfigure[End-effector orientation deviation from reference]{\includegraphics[width = .85\columnwidth]{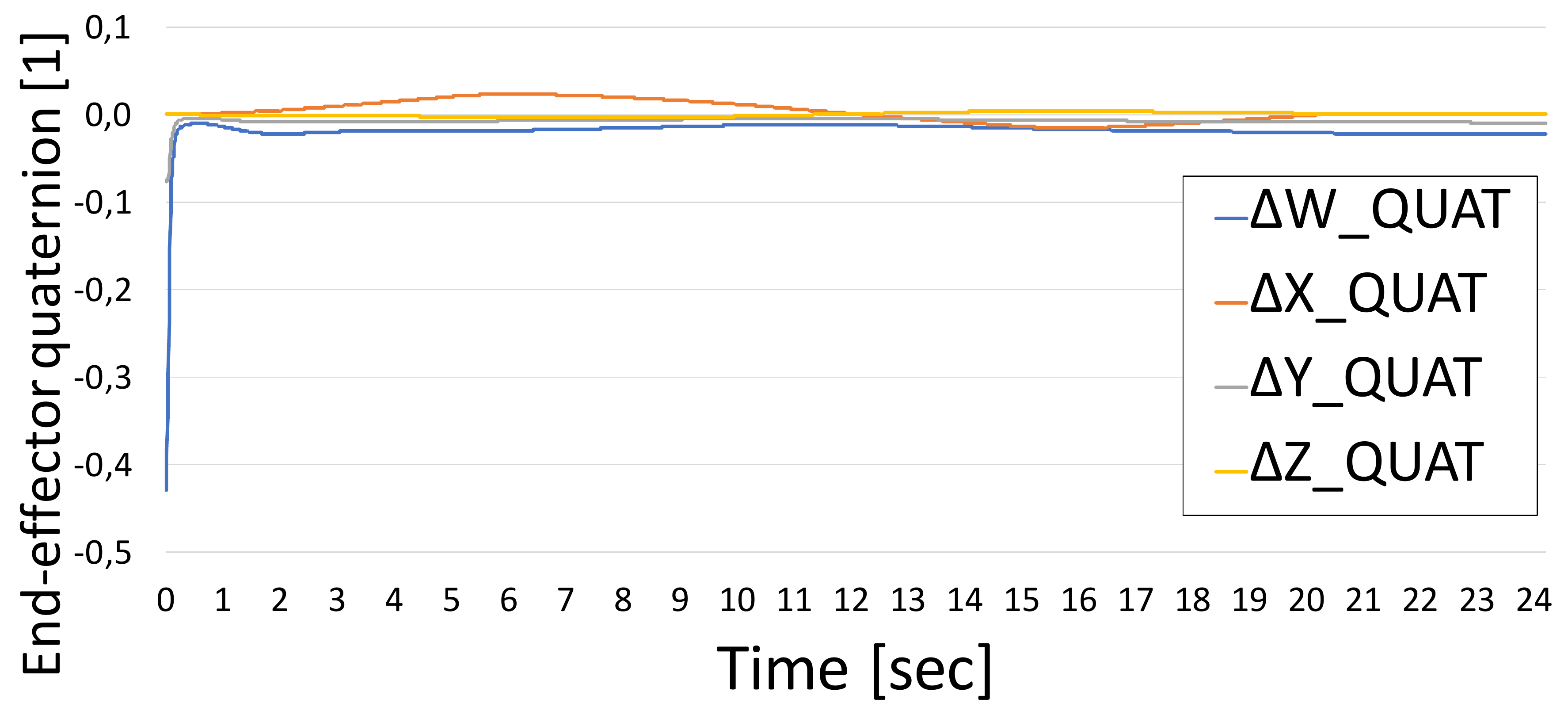}}
    \subfigure[OMAV orientation]{\includegraphics[width = .85\columnwidth]{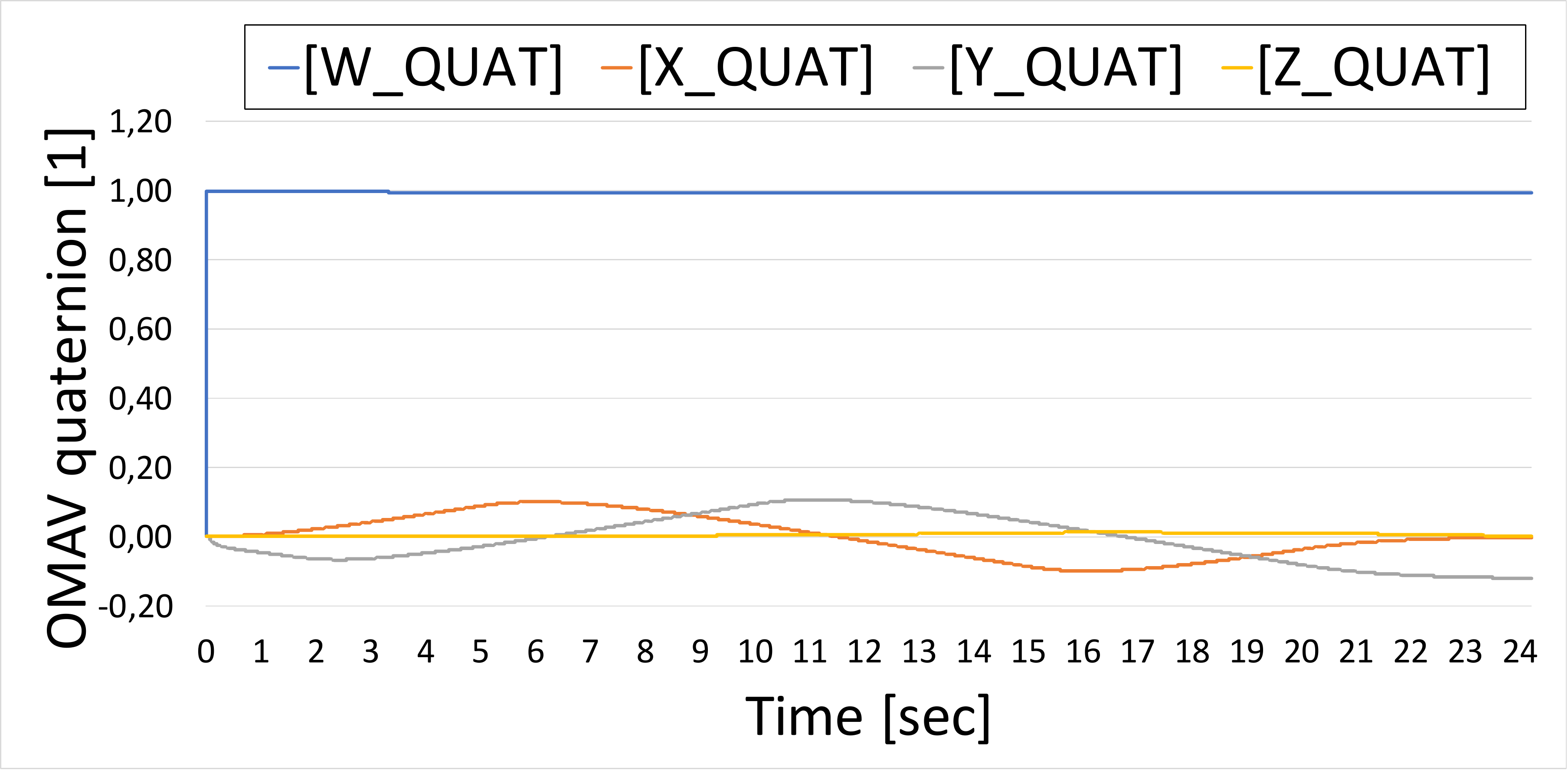}}
    \caption{Exploitation of nullspace motion. The end-effector is regulated and held at the constant point $[0, 0, -0.25]^T$\si{\meter} and the constant orientation of $15^{\circ}$ rotation around the $y$-axis with respect to the arm's straight configuration. The OMAV is constantly rotated around its axes. Subplot (a) shows the end-effector position. Subplot (b) shows the end-effector orientation in quaternion notation. Subplot (c) shows the orientation of the OMAV in quaternion notation. \label{fig:sim}}
    \vspace{-0.5cm}
\end{figure}

\begin{figure}[t]
	\centering
    \includegraphics[width = .9\columnwidth]{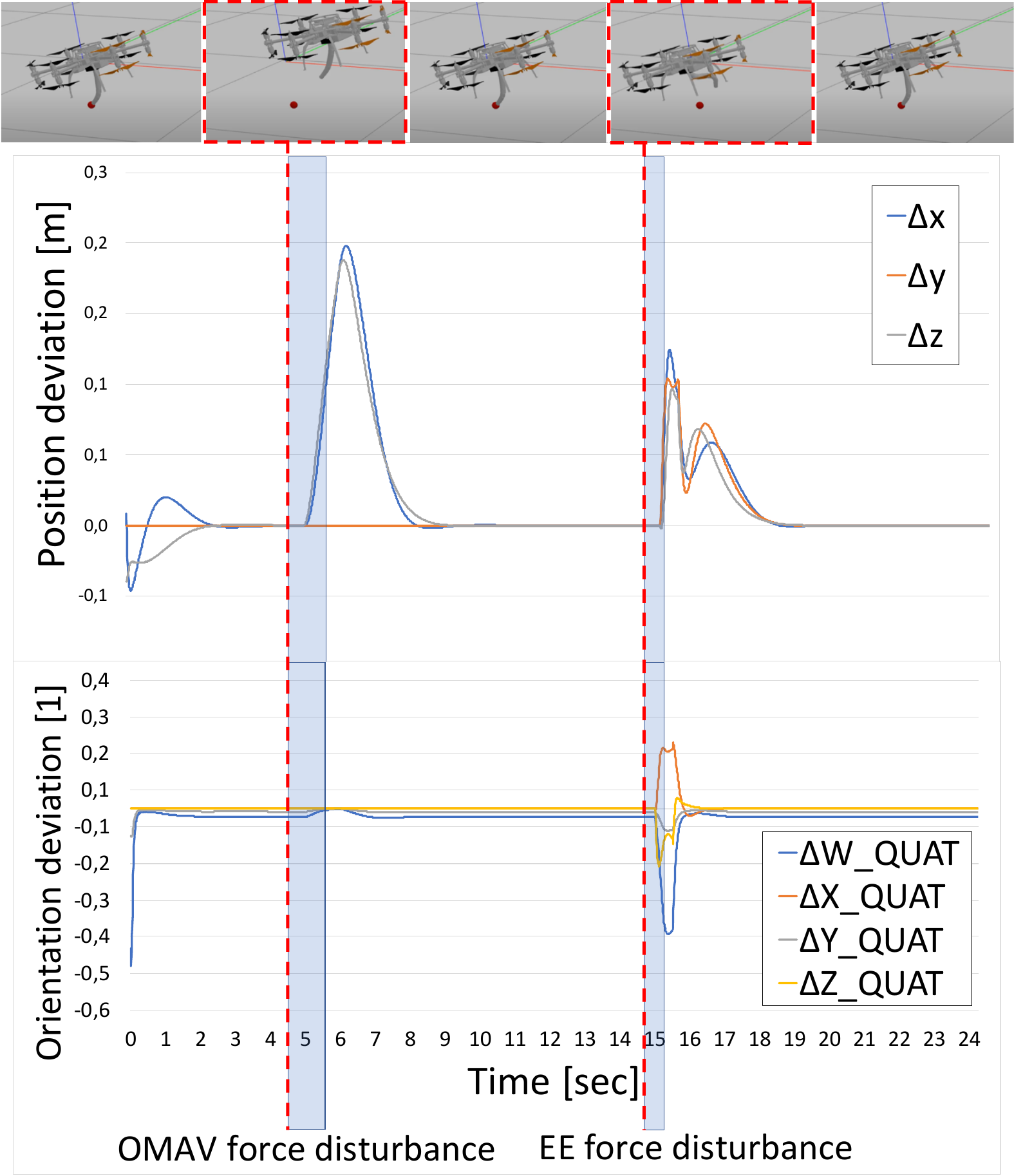}
    \caption{System evolution under force disturbances. The first disturbance of $[1,0,1]^T$\si{\newton} in frame $\mathcal{B}$ acts on the OMAV, occurs at $t=\SI{5}{\second}$, and lasts for \SI{1}{\second}. The second disturbance of $[1,1,1]^T$\si{\newton} in frame $\mathcal{E}$ acts directly on the end-effector, occurs at $t=\SI{15}{\second}$, and lasts for \SI{0.5}{\second}. The plots show the evolution of the position and orientation deviations of the end-effector. The top row of images above the plots shows the simulated system when experiencing these two disturbances.\label{fig:sim2}}
    \vspace{-0.5cm}
\end{figure}

\begin{figure*}[t]
	\centering
    \includegraphics[width = 1.65\columnwidth]{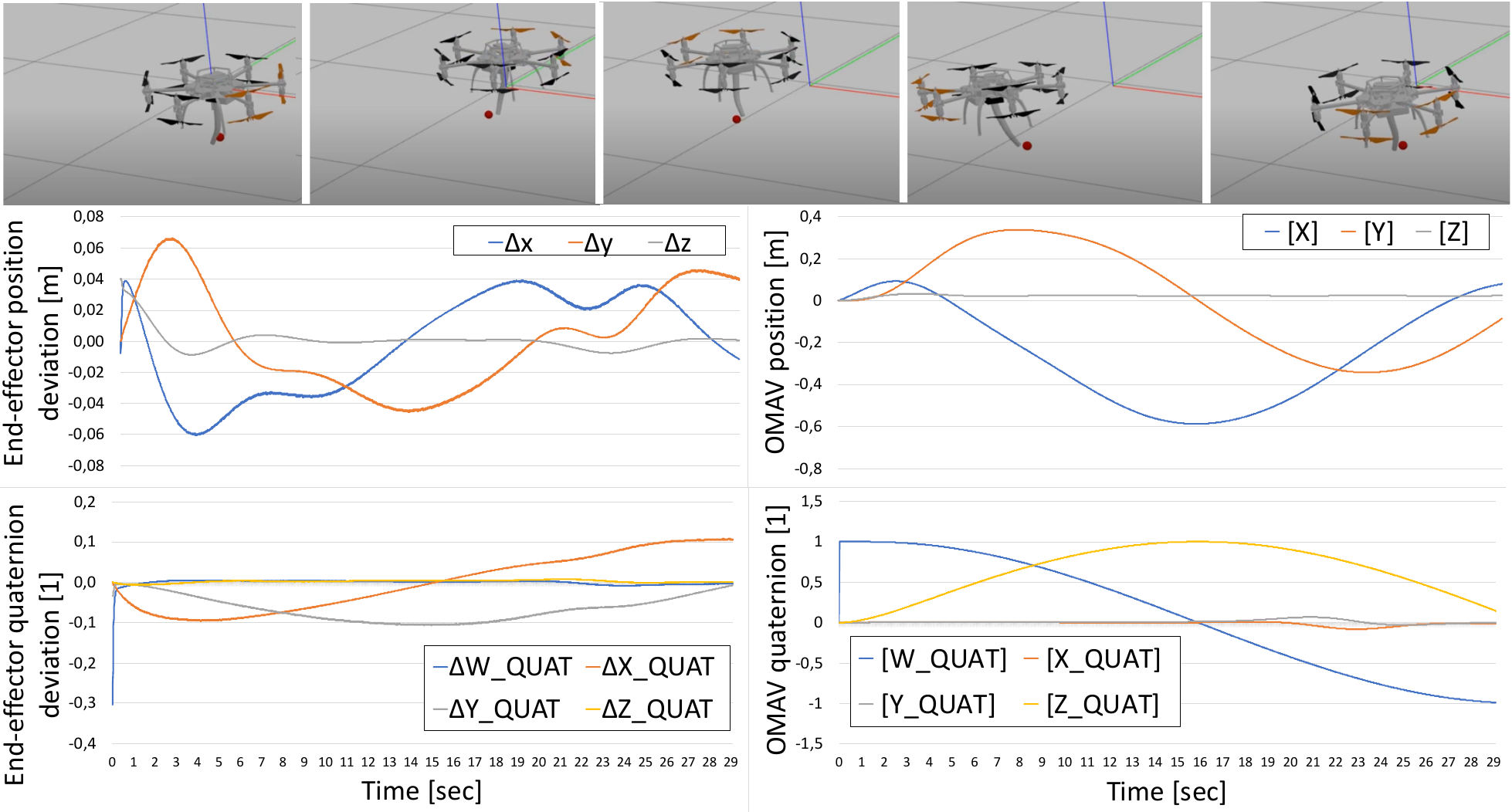}
    \caption{System evolution with the reference point moving in a circle of radius \SI{0.25}{\meter} at the center point $[-0.25,0,-0.25]^T \si{\meter}$. Both the OMAV and the soft arm's end-effector are tracking an always "inward" facing orientation. \label{fig:sim3}}
    \vspace{-0.5cm}
\end{figure*}

We use for the simulation a two-segment \emph{SoPrA} soft continuum arm~\cite{yasu2021sopra} coupled to an OMAV with six dual propellers. The mass (including the batteries) of the OMAV is $\SI[prefixes-as-symbols]{3.67}{\kilo\gram}$ and the moments of inertia are $\mathrm{diag}(0.075,0.073,0.139) \si{\kilo\gram\metre\squared}$.
The length of each SoPrA arm segment is $\SI[prefixes-as-symbols]{0.125}{\meter}$. The length of each connector piece between the segments is $\SI[prefixes-as-symbols]{0.02}{\meter}$ and is considered as an unactuated, rigid extension. The mass of the first segment is $\SI[prefixes-as-symbols]{0.190}{\kilo\gram}$, the second segment is $0.160kg$, and the connectors are $\SI[prefixes-as-symbols]{0.020}{\kilo\gram}$, each. The arm's diameter at the base is $\SI[prefixes-as-symbols]{0.042}{\meter}$, between the first and second segment is $\SI[prefixes-as-symbols]{0.035}{\meter}$, and at the tip is $\SI[prefixes-as-symbols]{0.028}{\meter}$. The remaining system properties were calculated as stated in~\cite{yasu2021sopra}.


The control gain and damping matrices were fine-tuned empirically and the final values can be seen in \Cref{table:matrices}. The tuple of three numbers in the table is interpreted as $\mathrm{diag}(*)$. $\bm{K}_*$ matrices have the dimension $\si{\newton\per\meter}$ and $\bm{D}_*$ are of $\si{\newton\second\per\meter}$. $\bm{K}_{EE_{pos}}$ and $\bm{D}_{EE_{pos}}$ are slightly different for the static regulation and dynamic trajectory tracking cases: since the accuracy in the regulation tasks is crucial, higher gains are proposed to ensure a fast, aggressive, and accurate enough control, whereas for motion tracking the safety (more compliance) and stability are the primary concerns leading to lower gains.

\subsection{Results}

The first experiment on \Cref{fig:sim} demonstrates the exploitation of the motion's nullspace. The OMAV was commanded to change its orientation around a given rotation axis by a certain amount of degrees $axis_{degree}$ in a continuous manner in the repeating cycle $y_{-15^{\circ}} \to x_{15^{\circ}} \to y_{15^{\circ}} \to x_{-15^{\circ}} \to y_{-15^{\circ}}$. Meanwhile, the position and orientation of the end-effector remained stable and close to the reference. After the transient behaviour, the end-effector orientation and position remain stable with negligible error to the reference, even though the OMAV was changing its orientation during the whole ($24 \si{\second}$) simulation.

In another experiment, the recovery capability after a disturbance was tested. The simulation shown in \Cref{fig:sim2} shows that the system is able to recover fast from disturbances acting either on the OMAV or on the soft arm. After the disturbance is applied, the system is able to react fast and returns to the reference position and orientation without any overshooting. These disturbances are not modeled, thus active disturbance rejection is not possible and not desired when collaborating with humans where a compliant behavior is preferred over a stiff one.

The third simulation examined the dynamic trajectory tracking capabilities of the controller shown in \Cref{fig:sim3}. While the system is capable of following the circular trajectory with a given orientation over time, there was a small lag introduced in the position tracking. The mean L2 norm of the position error is around $0.043m$. This is due to the fact that the control is governed by the $\Delta$ offset terms introduced in \Cref{fig:blockdiagram}, which are arbitrary small at vanishing differences between the reference and control variable. It can be understood as follows: the controller needs a certain ``minimal'' difference between the reference and the controlled value to be ``active'' and effective enough. Note that we did not employ any model predictive control techniques.

\section{Conclusion and future work}
In this paper we propose a unified mathematical description of a coupled flying system consisting of an OMAV and a soft continuum arm. We show how extending the floating base approach taken from the classical rigid-body robotics literature leads to an analogous formulation in the PCC space with the ARBM. Furthermore, we derive a hierarchical task-prioritisation control architecture tailored to the coupled system. The approach has been tested and evaluated in various simulation scenarios. The system's architecture exhibits certain desirable characteristics, such as the exploration of nullspace-motion, disturbance recovery, and trajectory tracking with a given orientation of the end-effector and the OMAV.

We hope that with this work we can lay the foundation for future research in the area of soft continuum manipulators mounted to flying vehicles. We believe that their potential for industrial applications is tremendous in the ever-growing demand on semi-automated warehouses, where humans and robots would actively and efficiently collaborate to retrieve and disperse the goods. Future work will focus on conducting experiments on the real-world system, potentially with an additional gripper attached at the end-effector, to test some basic transport capabilities. Another focus will be placed on micro oscillation effects observed during control simulations using high gains. These effects could be addressed using either curvature space or hybrid control, where the gains for the OMAV and SoPrA can be adjusted independently.
%
%



%
%







\section*{ACKNOWLEDGMENT}
We are grateful for Ing. Róbert Szász Sr.'s support in editing the video material accompanying this paper. 


\bibliographystyle{IEEEtran}
\bibliography{IEEEabrv,09_references}

\begin{thebibliography}{10}
\providecommand{\url}[1]{#1}
\csname url@samestyle\endcsname
\providecommand{\newblock}{\relax}
\providecommand{\bibinfo}[2]{#2}
\providecommand{\BIBentrySTDinterwordspacing}{\spaceskip=0pt\relax}
\providecommand{\BIBentryALTinterwordstretchfactor}{4}
\providecommand{\BIBentryALTinterwordspacing}{\spaceskip=\fontdimen2\font plus
\BIBentryALTinterwordstretchfactor\fontdimen3\font minus
  \fontdimen4\font\relax}
\providecommand{\BIBforeignlanguage}[2]{{%
\expandafter\ifx\csname l@#1\endcsname\relax
\typeout{** WARNING: IEEEtran.bst: No hyphenation pattern has been}%
\typeout{** loaded for the language `#1'. Using the pattern for}%
\typeout{** the default language instead.}%
\else
\language=\csname l@#1\endcsname
\fi
#2}}
\providecommand{\BIBdecl}{\relax}
\BIBdecl

\bibitem{2021-BodTogSie}
K.~Bodie, M.~Tognon, and R.~Siegwart, ``Dynamic end effector tracking with an
  omnidirectional parallel aerial manipulator,'' \emph{IEEE Robotics and
  Automation Letters}, vol.~6, no.~4, pp. 8165--8172, 2021.

\bibitem{2021g-OllTogSuaLeeFra}
A.~Ollero, M.~Tognon, A.~Suarez, D.~J. Lee, and A.~Franchi, ``Past, present,
  and future of aerial robotic manipulators,'' \emph{IEEE Trans. on Robotics},
  2021.

\bibitem{allenspach_OMAV}
M.~Allenspach, K.~Bodie, M.~Brunner, L.~Rinsoz, Z.~Taylor, M.~Kamel,
  R.~Siegwart, and J.~Nieto, ``\BIBforeignlanguage{en}{Design and optimal
  control of a tiltrotor micro-aerial vehicle for efficient omnidirectional
  flight},'' \emph{\BIBforeignlanguage{en}{The International Journal of
  Robotics Research 1–21}}, pp. 1--21, March 2020.

\bibitem{hwisu_compliance_joint}
K.~Hwi-Su, K.~In-Moon, C.~Chang-Nho, and S.~Jae-Bok, ``{Safe joint module for
  safe robot arm based on passive and active compliance method},''
  \emph{Mechatronics}, vol.~22, pp. 1023--1030, 2012.

\bibitem{dissertation_compliant_links}
Y.~She, ``Compliant robotic arms for inherently safe physical human-robot
  interaction,'' Ph.D. dissertation, The Ohio State University, 2018.

\bibitem{schumacher2019_compliant_control}
M.~Schumacher, J.~Wojtusch, P.~Beckerle, and O.~Von~Stryk, ``{An Introductory
  Review of Active Compliant Control},'' \emph{Robotics and Autonomous
  Systems}, 2019.

\bibitem{hutter_starleth_compliance_actuator}
M.~Hutter, C.~Remy, M.~A. Hoepflinger, and R.~Siegwart, ``{High Compliant
  Series Elastic Actuation for the Robotic Leg ScarlETH},'' \emph{International
  Conference on Climbing and Walking Robots and the Support Technologies for
  Mobile Machines (CLAWAR)}, vol.~14, no.~1, pp. 507--514, 2011.

\bibitem{hutter_starleth_compliance_control}
C.~Gehring, S.~Coros, M.~Hutter, C.~D. Bellicoso, H.~Heijnen, R.~Diethelm,
  M.~Bloesch, P.~Fankhauser, J.~Hwangbo, M.~A. Hoepflinger, and R.~Siegwart,
  ``{Practice Makes Perfect: An Optimization-Based Approach to Controlling
  Agile Motions for a Quadruped Robot},'' \emph{IEEE Robotics \& Automation
  Magazine}, vol.~23, no.~1, pp. 34--43, 2016.

\bibitem{katzschmann2015softarm2D}
R.~K. Katzschmann, A.~D. Marchese, and D.~Rus, ``{Autonomous Object
  Manipulation Using a Soft Planar Grasping Manipulator},'' \emph{Soft
  Robotics}, vol.~2, no.~4, pp. 155--164, dec 2015.

\bibitem{katzschmann2020softarm2Dcontrol}
C.~Della~Santina, R.~K. Katzschmann, A.~Bicchi, and D.~Rus, ``{Model-based
  dynamic feedback control of a planar soft robot: trajectory tracking and
  interaction with the environment},'' \emph{The International Journal of
  Robotics Research}, 2020.

\bibitem{katzschmann2019softarm3Dcontrol}
R.~K. Katzschmann, C.~Della~Santina, Y.~Toshimitsu, A.~Bicchi, and D.~Rus,
  ``{Dynamic Motion Control of Multi-Segment Soft Robots Using Piecewise
  Constant Curvature Matched with an Augmented Rigid Body Model},'' \emph{IEEE
  International Conference on Soft Robotics (RoboSoft)}, 2019.

\bibitem{katzschmann2019softarm3DFEM}
R.~K. Katzschmann, M.~Thieffry, O.~Goury, A.~Kruszewski, T.-M. Guerra,
  C.~Duriez, and D.~Rus, ``{Dynamically Closed-Loop Controlled Soft Robotic Arm
  using a Reduced Order Finite Element Model with State Observer},'' \emph{IEEE
  International Conference on Soft Robotics (RoboSoft)}, 2019.

\bibitem{2021f-HamUsaSabStaTogFra}
M.~Hamandi, F.~Usai, Q.~Sable, N.~Staub, M.~Tognon, and A.~Franchi, ``Design of
  multirotor aerial vehicles: a taxonomy based on input allocation,'' \emph{The
  International Journal of Robotics Research}, vol.~40, no. 8-9, pp.
  1015--1044, 2021.

\bibitem{sadati2018control}
S.~H. Sadati, S.~E. Naghibi, I.~D. Walker, K.~Althoefer, and T.~Nanayakkara,
  ``Control space reduction and real-time accurate modeling of continuum
  manipulators using ritz and ritz--galerkin methods,'' \emph{IEEE Robotics and
  Automation Letters}, vol.~3, no.~1, pp. 328--335, 2018.

\bibitem{renda2017discrete}
F.~Renda, F.~Boyer, J.~Dias, and L.~Seneviratne, ``Discrete cosserat approach
  for multi-section soft robots dynamics,'' \emph{arXiv preprint
  arXiv:1702.03660}, 2017.

\bibitem{dandrea_omav}
D.~Brescianini and R.~D'Andrea, ``\BIBforeignlanguage{en}{Design, modeling and
  control of an omni-directional aerial vehicle},''
  \emph{\BIBforeignlanguage{en}{2016 IEEE International Conference on Robotics
  and Automation (ICRA)}}, pp. 1--6, May 2016.

\bibitem{ryll_omav}
M.~Ryll, H.~H. Bülthoff, and P.~R. Giordano, ``\BIBforeignlanguage{en}{A novel
  overactuated quadrotor uav: Modeling,control and experimental validation},''
  \emph{\BIBforeignlanguage{en}{IEEE Transactions on Control Sys-tems
  Technology, Institute of Electrical and Electronics Engineers}}, pp. 1--10,
  January 2015.

\bibitem{siegwart_OMAV}
K.~Bodie, Z.~J. Taylor, M.~S. Kamel, and R.~Siegwart,
  ``\BIBforeignlanguage{en}{Towards efficient full pose omnidirectionality with
  overactuated mavs},'' \emph{\BIBforeignlanguage{en}{Proceedings of the 2018
  International Symposium on Experimental Robotics (pp.85-95)}}, pp. 1--10,
  January 2020.

\bibitem{siegwart_OMAV_force_control}
K.~Bodie, M.~Brunner, M.~Pantic, S.~Walser, P.~Pfändler, U.~Angst, J.~Nieto,
  and R.~Siegwart, ``\BIBforeignlanguage{en}{Active interaction force control
  for omnidirectional aerial contact-based inspection},''
  \emph{\BIBforeignlanguage{en}{IEEE Transactions on Robotics}}, vol.~37, pp.
  709--722, December 2020.

\bibitem{2019e-TogTelGasSabBicMalLanSanRevCorFra}
M.~Tognon, H.~A. {Tello~Ch\'avez}, E.~Gasparin, Q.~Sabl\'e, D.~Bicego,
  A.~Mallet, M.~Lany, G.~Santi, B.~Revaz, J.~Cort\'es, and A.~Franchi, ``A
  truly redundant aerial manipulator system with application to push-and-slide
  inspection in industrial plants,'' \emph{IEEE Robotics and Automation
  Letters}, vol.~4, no.~2, pp. 1846--1851, 2019.

\bibitem{yasu2021sopra}
Y.~Toshimitsu, K.~W. Wong, T.~Buchner, and R.~K. Katzschmann, ``{SoPrA:
  Fabrication \& Dynamical Modeling of a Scalable Soft Continuum Robotic Arm
  with Integrated Proprioceptive Sensing},'' \emph{2021 IEEE/RSJ International
  Conference on Intelligent Robots and Systems (IROS 2021)}, 2021.

\bibitem{kannan_control_uav_operational_space}
S.~Kannan, Q.~Guzman, J.~Dentler, M.~Olivares-Mendez, and H.~Voos,
  ``\BIBforeignlanguage{en}{Control of aerial manipulation vehicle in
  operational space},'' \emph{\BIBforeignlanguage{en}{2016 8th International
  Conference on Electronics, Computers and Artificial Intelligence (ECAI)}},
  pp. 1--5, June 2016.

\bibitem{caccavale_control_uav_robot_arm}
F.~Caccavale, G.~Giglio, G.~Muscio, and F.~Pierri,
  ``\BIBforeignlanguage{en}{Adaptive control for uavs equipped with a robotic
  arm},'' \emph{\BIBforeignlanguage{en}{2014 19th World Congress (IFAC)}}, pp.
  1--6, August 2014.

\bibitem{fishman2021softdrone}
J.~Fishman, S.~Ubellacker, N.~Hughes, and L.~Carlone, ``{Dynamic Grasping with
  a "Soft" Drone: From Theory to Practice},'' \emph{2021 IEEE/RSJ International
  Conference on Intelligent Robots and Systems (IROS 2021)}, 2021.

\bibitem{nakanishi2007floatingbase}
J.~Nakanishi, M.~Mistry, and S.~Schaal, ``{Inverse Dynamics Control with
  Floating Base and Constraints},'' \emph{IEEE International Conference on
  Robotics and Automation}, 2007.

\bibitem{katzschmann2018softarm2Dcontrol}
C.~Della~Santina, R.~K. Katzschmann, A.~Bicchi, and D.~Rus, ``{Dynamic Control
  of Soft Robots Interacting with the Environment},'' \emph{2018 IEEE
  International Conference on Soft Robotics (RoboSoft)}, 2018.

\bibitem{robot_dynamics_lecture_notes_eth}
\BIBentryALTinterwordspacing
E.~Z. Robotic Systems~Lab, \emph{Robot Dynamics Lecture Notes}.\hskip 1em plus
  0.5em minus 0.4em\relax ETH Zürich, 2017. [Online]. Available:
  \url{https://ethz.ch/content/dam/ethz/special-interest/mavt/robotics-n-intelligent-systems/rsl-dam/documents/RobotDynamics2017/RD_HS2017script.pdf}
\BIBentrySTDinterwordspacing

\bibitem{dietrich2015nullspace_projection}
A.~Dietrich, C.~Ott, and A.~Albu-Schäffer, ``{An overview of null space
  projections for redundant, torque-controlled robots},'' \emph{The
  International Journal of Robotics Research (IJRR)}, 2015.

\bibitem{ros}
\BIBentryALTinterwordspacing
{Stanford Artificial Intelligence Laboratory et al.}, ``Robotic operating
  system.'' [Online]. Available: \url{https://www.ros.org}
\BIBentrySTDinterwordspacing

\bibitem{drake}
\BIBentryALTinterwordspacing
R.~Tedrake and the Drake Development~Team, ``Drake: Model-based design and
  verification for robotics,'' 2019. [Online]. Available:
  \url{https://drake.mit.edu}
\BIBentrySTDinterwordspacing

\end{thebibliography}

\end{document}